\pdfoutput=1

\documentclass[11pt]{article}
\usepackage{wasysym}

\usepackage[preprint]{acl}

\usepackage{times}
\usepackage{latexsym}
\usepackage{diagbox}
\usepackage{verbatim}
\usepackage{wrapfig}
\usepackage{longtable}

\usepackage[T1]{fontenc}

\usepackage[utf8]{inputenc}

\usepackage{microtype}

\usepackage{inconsolata}

\usepackage{graphicx}

\newcommand{\ag}{\textcolor{violet}{Ag}}
\newcommand{\pat}{\textcolor{orange}{Pat}}
\newcommand{\akt}{\textcolor{brown}{Akt}}
\newcommand{\pass}{\textcolor{teal}{Pass}}
\newcommand{\prep}{\textcolor{blue}{p}}
\newcommand{\by}{\textcolor{red}{by}}

\title{
Generalization through Lexical Abstraction in Transformer Models: \\ The Case of Functional Words
}

\author{
  \textbf{Giuseppe Samo\textsuperscript{1}} \and
  \textbf{Vivi Nastase\textsuperscript{1}} \and
  \textbf{Paola Merlo\textsuperscript{1,2}}
\\
  \textsuperscript{1}Idiap Research Institute, Martigny, Switzerland \\
  \textsuperscript{2}University of Geneva, Swizerland
  \\
\texttt{giuseppe.samo@idiap.ch, vivi.a.nastase@gmail.com, Paola.Merlo@unige.ch}
}

\begin{document}
\maketitle
\begin{abstract}


Pronouns, adverbs and other functional words (such as \textit{they, her, somewhere, there}) are often used in language to replace concrete nouns or phrases, when their properties -- such as gender, grammatical number -- provide sufficient information for the given context. Do pretrained transformer models encode such functional words in a manner that allows them to be used like humans do? Can language models recognize the syntactic and semantic parallelism of sentences such as "The researchers wrote the paper" and "They wrote it", which relies on such lexical abstraction? 

We map these linguistic questions into the embedding space of a pretrained transformer model, and compare representations of nouns, with the representations of the pronouns and adverbs that can replace these nouns, in isolation and in parallel lexicalized and functional sentences. We then probe for shared syntactic and semantic structure in the embeddings of parallel lexicalized and functional sentences. 

We find that functional words are located centrally compared to nouns, but are also distinct, which is congruent with their behaviour as place-holders in a wide variety of contexts. 
The analysis of the embeddings of parallel (lexicalized and functional) sentences show them inhabiting different subspaces of the embedding space. Experiments that distil the structural information of the sentence show that training on either type of data does not reveal the shared structure -- because of the over-consistency of the vocabulary (in case of the functional data), and the too much variety (in case of the lexicalized versions). However, training with a mix of functional and lexicalized sentences, the shared structure emerges.

\end{abstract}

\section{Introduction}

Foundational language models are very successful, and much of their success stems from their ability to induce word or token representations that encode the extremely complex language data, with many generative factors \cite{bengio-ea2013}. It is an ongoing quest to understand these representations, the kind of linguistic information they encode and the way a system is able to successfully manipulate them to solve a wide variety of tasks. It is difficult to attribute their high performance on numerous linguistic and NLP tasks to the language models' understanding of language and its structure \cite{waldis-etal-2024-holmes}. One of the criteria for judging the degree of language understanding in a model is their capacity to ``generalize" well. This question is often approached from a technical, rather than a linguistic, perspective. Generalization is considered a crucial property of a learned model, as it ensures trust in its deployment outside of its training environment -- whether this application involves a slightly different task, out-of-distribution data, a different language, or some other level of distinction between the application domain and the one it was trained on \cite{Hupkes2023}. 

We focus here on linguistic generalizations and abstractions. For example, speakers can easily strip down a sentence to a basic syntactic-semantic structure, such as \textit{Who did what to whom} or \textit{She put that there} or \textit{She does that sometimes.}  The use of pronouns or adverbs to reduce a sentence to a ``skeleton'' does not rely on using out-of-vocabulary items, as pronouns and adverbials, such as {\it somewhere/sometime}, are some of the most frequent words in a corpus, and appear in many shared contexts, as their frequent use in coreferring expressions attests.
In semantics, pronominal forms are usually treated as variables, placeholders for more structured lexical elements within a sentence and thus highly abstract entities \citep{Büring+2019+1+32}. 

Is this particular property of functional words -- as abstract place-holders for nouns and prepositional phrases -- captured in language models? We map this question into the embedding space of a pretrained transformer model (PTM). Embedding spaces are built on the assumption that similarity and relatedness between words is equated with closeness in the embedding space. As abstract place-holders, functional words appear in shared contexts with a wide variety of nouns and prepositional phrases, thus they are similar to a certain degree. We expect them to be somewhere in the center of the embedding space, so they can be close to a wide variety of nouns and prepositional phrases. In the sentence embedding space, we test whether parallel sentences -- such as "The researchers wrote the paper" and "They wrote it" -- are encoded with embeddings that are close in space as they share syntactic (the sentence structure and phrase types) and semantic properties (semantic roles, and specific properties of the role fillers). If a PTM captures linguistic abstraction, this shared information -- syntactic structure and semantic roles -- should be encoded in a similar manner regardless of whether the phrases contain nouns or functional words.  

We test these expectations via purposefully generated dataset on the causative verb alternation, with structure and lexical variation at multiple levels. We extract data in several formats, to explore words in isolation and in context, and the functional and lexicalized versions of sentences. We observe that functional words are rather central in the embedding space compared to nouns, but also isolated, which matches the expectations arising from their behaviour as place-holders in a wide variety of contexts. Analyses of sentence embeddings show that the functional and lexicalized versions inhabit different areas in the embedding space, which, according to the commonly used geometry, with the cosine as a distance metric, would indicate that the lexicalized and functional versions of a sentence do not have much in common. If we consider, however, sentence embeddings to be complex objects, we can detect their shared syntactic-semantic structure, encoded in a similar manner.

\section{Data}
\label{sec:data}

Verb alternations require observing at least two related sentences. They show that the same verb can appear in different sentential contexts, with systematic syntactic-semantic mappings of their arguments across the sentences, like a system of equations that all share the same variable bindings. 

The dataset \citep{samo2025-structured} is generated from a set of verbs belonging to the change-of-state (COS) and object-drop (OD) classes \citep{levin93}. These classes provide an argument structure minimal pair: they share the same syntactic structure - transitive/intransitive alternation - but differ in their 
argument structure. The object of the transitive verbs belonging to the COS class bears the same semantic role (Patient) as the subject of the intransitive verb (\textit{The artist opens this door/This door opens}). The transitive form of the verb has a causative meaning. In contrast, for OD verbs the subject bears the same semantic role (Agent) in both the transitive and intransitive forms and the verb does not have a causative meaning (\textit{The artist paints this door/The artist paints}) \citep{levin93,merlo2001automatic}. 

We divide words into lexical and functional. Lexical elements, or content words, are an open class of words with a meaningful content, corresponding to concepts or entities and events in the world. The role of the closed class of function words, instead, is to express grammatical functions. We focus specifically on pronouns, and a subset of adverbs, those  that can express temporal and spatial concepts. These function words  can be used as general placeholders for nouns and prepositional phrases: for instance, \textit{The researchers wrote the article last week} can also be expressed more abstractly as  \textit{They wrote it then.}

\subsection{Data templates}

The dataset comprises instances that follow the Blackbird Language Matrices framework \cite{merlo2023}. Each instance is a multiple-choice puzzle and it 
consists of (i) a rule-generated \textit{context} sequence of sentences that illustrate the encoded phenomenon. The rules are of two types: rules that described the linguistic property under study (verb alternation) and rules that are not related to it (e.g. presence or absence of a prepositional phrase). One sentence that would make the sequence complete is missing, and must be chosen from (ii) an \textit{answer set} of minimally differing contrastive sentences -- one correct, and each of the others violating a sub-rule.

\paragraph{Context set}

The syntax-semantics features of the verb alternation, and their combination rules, lead to the construction of the context set.  Specifically, (i) the presence of one or two arguments and their attributes (agents, \ag; patients, \pat) ; (ii) the active (\akt) or passive (\pass) voice of the verb. The phenomenon-external factors include an alternation  between a NP introduced (i) by any preposition (e.g., \textit{in an instant}, henceforth \prep-NP) and (ii) by the preposition by (e.g., \textit{by chance}, \by-NP), but not agentive (e.g., \textit{by the artist}, \by-\ag/\by-\pat), which remains a confounding variable.
The OD context minimally differs from the COS in the last sentence of the context: the subject of the intransitive is an Agent, and not the Patient. 

\paragraph{Answer Set}


All answers have the same structure: (NP V \by-NP) consisting of a verb, two nominal constituents (giving rise to a structure of the type NP V NP) and a preposition (\by, or the lack of the preposition) between the verb and the second NP. 
The candidate answers comprise the correct intransitive form of the alternation followed by a \by-NP which satisfy the rules of the BLM, and the contrastive incorrect answers obtained by corrupting some properties of the rules (wrong argument, wrong voice of the verb, lack of preposition, wrong nominal constituent of the PP).\footnote{Error types: wrong semantic role on the first constituent is a syntax-semantic mapping error (SSM), wrong last constituent introduced by the preposition \it{by} \textsc{WrBy}, and the other errors are labelled according to the type of resulting structure -- intransitive, \textsc{Intr}; transitive, \textsc{Trans}; passive, \textsc{Pass}.} 

The answer set does not change across verb classes, only the label of the correct answer: the correct  answer for COS is an error for OD, and viceversa.  
The BLM-template (context and answers) for COS and OD are presented in Figure~\ref{tab:templateBLM}.


\begin{figure}[!h]
\centering
\scriptsize 
\setlength{\tabcolsep}{2pt} 

\begin{tabular}{lllll} 
\hline
\multicolumn{5}{c}{\sc COS context}\\
\hline
1 & \ag & \akt & \pat & \prep-NP \\
2 & \ag & \akt & \pat  & \by-NP  \\
3 & \pat & \pass & \by-\ag & \prep-NP \\
4 & \pat & \pass & \by-\ag & \by-NP \\
5 & \pat & \pass & & \prep-NP \\
6 & \pat & \pass & & \by-NP \\
7 & \pat & \akt & & \prep-NP \\ 
8 & ??? & & &  \\ 
\end{tabular}
\hspace{0.5cm} 
\begin{tabular}{ll|l} \hline
\multicolumn{3}{c}{\sc COS answer set}  \\ \hline
1 & \textcolor{orange}{Pat} \textcolor{brown}{Akt}   \textcolor{red}{by}-NP & \textsc{\textbf{Correct}}\\ 
2 & \textcolor{violet}{Ag} \textcolor{brown}{Akt}   \textcolor{red}{by}-NP & \textsc{SSM-Int}  \\ 
3 & \textcolor{orange}{Pat} \textcolor{teal}{Pass}  \textcolor{red}{by}-\textcolor{violet}{Ag} & \textsc{Pass}\\
4 & \textcolor{violet}{Ag} \textcolor{teal}{Pass}   \textcolor{red}{by}-\textcolor{orange}{Pat} & \textsc{SSM-Pass}\\
5 & \textcolor{orange}{Pat} \textcolor{brown}{Akt}   \textcolor{violet}{Ag} & \textsc{Trans}\\
6 & \textcolor{violet}{Ag} \textcolor{brown}{Akt}   \textcolor{orange}{Pat} & \textsc{Ssm-Trans} \\
7 & \textcolor{orange}{Pat} \textcolor{brown}{Akt}   \textcolor{red}{by}-\textcolor{violet}{Ag} & \textsc{WrBy} \\
8 & \textcolor{violet}{Ag} \textcolor{brown}{Akt}   \textcolor{red}{by}-\textcolor{orange}{Pat} & \textsc{SSM-WrBy}  \\ 
\end{tabular}

\vspace{2mm}
\vspace{2mm}

\begin{tabular}{lllll} 
\hline
\multicolumn{5}{c}{\sc OD context}\\
\hline
1 & \ag & \akt & \pat & \prep-NP\\
2 & \ag & \akt & \pat & \by-NP \\
3 & \pat & \pass & \by-\ag  & \prep-NP\\
4 & \pat & \pass & \by-\ag  & \by-NP\\
5 & \pat & \pass & & \prep-NP\\
6 & \pat & \pass & & \by-NP\\
7 & \ag & \akt & & \prep-NP\\
8 & ??? & & &  \\
\end{tabular}
\hspace{0.5cm} 
\begin{tabular}{rl|l} \hline
\multicolumn{3}{c}{\sc OD answer set}  \\ \hline
1 & \textcolor{orange}{Pat} \textcolor{brown}{Akt}   \textcolor{red}{by}-NP & \textsc{SSM-Int}\\
2 & \textcolor{violet}{Ag} \textcolor{brown}{Akt}   \textcolor{red}{by}-NP & \textsc{\textbf{Correct}}  \\ 
3 & \textcolor{orange}{Pat} \textcolor{teal}{Pass}  \textcolor{red}{by}-\textcolor{violet}{Ag} & \textsc{SSM-Pass}\\
4 & \textcolor{violet}{Ag} \textcolor{teal}{Pass}   \textcolor{red}{by}-\textcolor{orange}{Pat} & \textsc{Pass}\\
5 & \textcolor{orange}{Pat} \textcolor{brown}{Akt}   \textcolor{violet}{Ag} & \textsc{SSM-Trans}\\
6 & \textcolor{violet}{Ag} \textcolor{brown}{Akt}   \textcolor{orange}{Pat} & \textsc{Trans} \\
7 & \textcolor{orange}{Pat} \textcolor{brown}{Akt}   \textcolor{red}{by}-\textcolor{violet}{Ag} & \textsc{SSM-WrBy} \\
8 & \textcolor{violet}{Ag} \textcolor{brown}{Akt}   \textcolor{red}{by}-\textcolor{orange}{Pat} & \textsc{WrBy}  \\ 

\end{tabular}

\caption{ BLM COS and OD contexts and answers.
}
\label{tab:templateBLM}
\end{figure}

\subsection{Levels of lexical abstraction}

\begin{figure}
    \centering
    \includegraphics[width=0.95\linewidth]{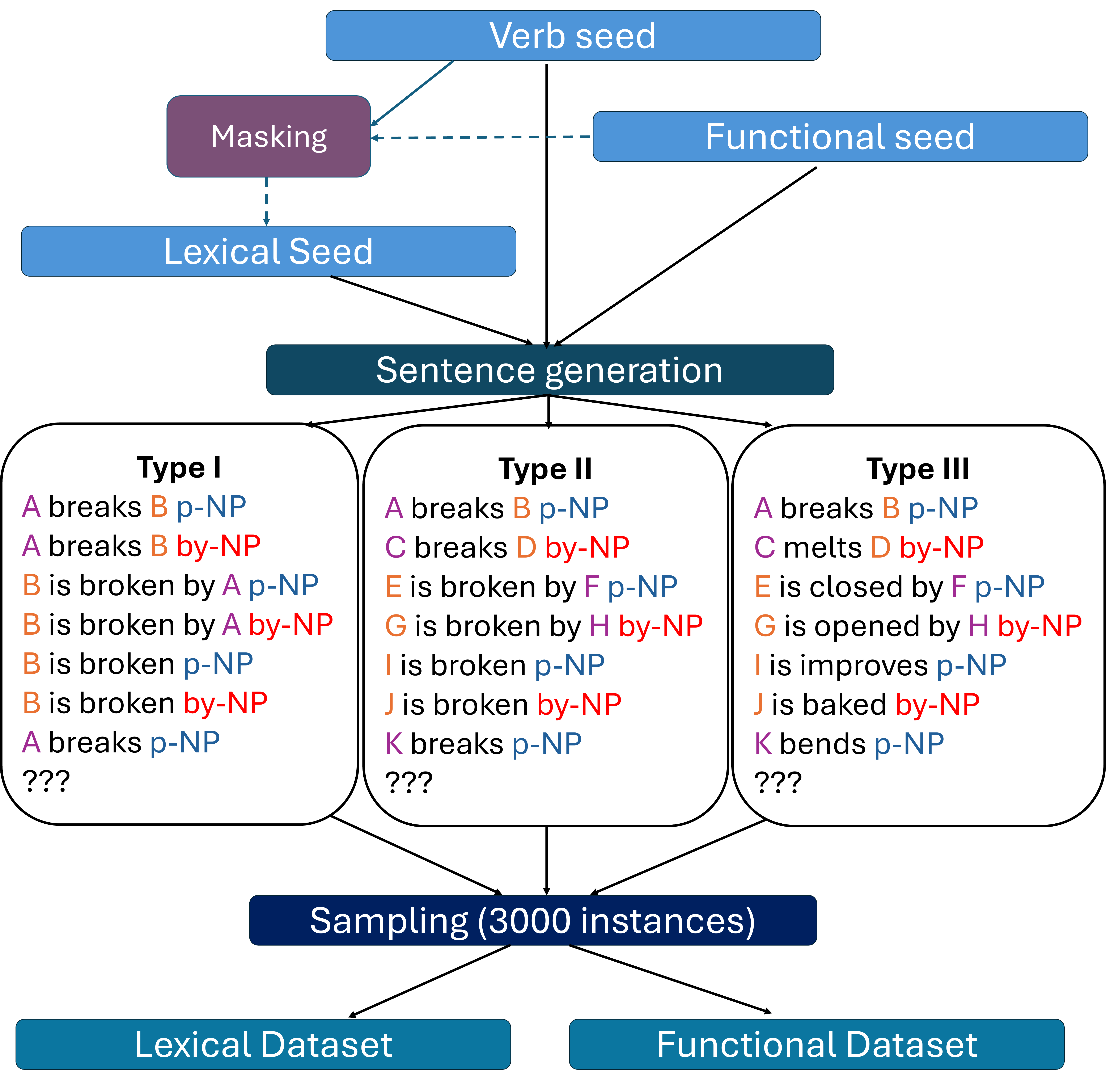}
    \caption{Process of generation of the three levels of lexical variation (type I, II, III), exemplified for COS data. Type I data contains instances with lexically consistent material, with minimal change across the context and the answer set. In type II the verb remains the same while one constituent varies across the context and the answer set. Type III data displays maximal lexical variation in both the context sentences and the answer set.}
    \label{fig:generation}
\end{figure}

To explore generalisation through abstraction, we produce two main  variants of the data -- a lexicalized one (labelled \textit{Lex}\/), and a functional one, where functional words replace all content words except the main verb (labelled \textit{Fun}\/).
The lexicalised variant comes in different types (type I, II, III), with varying amounts of lexicalisation, for  comparison with the small size inventory of the functional words.  The groups are exemplified in Figure \ref{fig:generation}, together with the generation process presented in the next paragraph. Figure \ref{tab:TypeI-COSfunlex} and Figure \ref{tab:TypeI-ODfunlex} in the appendix provide examples for type I data for both verb classes.  

\subsection{Main Dataset}

The main dataset is built based on thirty (manually chosen) verbs from each of the two classes discussed in \citet{levin93}. See Table \ref{tab:verbs-app} in the appendix for the full list.

The functional lexicon has been manually selected by the authors to maintain the syntactic and semantic acceptability of the sentences\footnote{Following the discussion in \citet{haspelmath1997indefinite}, we add elements like \textit{somebody} as pronominal elements.}. The lexical alternatives were provided by a masked language model (\textit{bert-base-uncased}, \citealt{devlin19}).
The models received sentences containing only the masked constituent, the verb and the functional elements. 
For example, to retrieve the arguments for the verb \textit{break}, two masked templates are used: the patient is masked and the agent is in pronominal form (e.g. \textit{she broke (the/a/some/...) \texttt{<MASK>}}), and  the subject of the transitive is masked and the patient is a pronoun \textit{(e.g. (the/a/some/...) \texttt{<MASK>} broke it}. 
Both the lexical seed and the functional seed contain five semantically plausible instances for each constituent class (\ag, \pat, \prep-NP and \by-NP). We ensured a balanced distribution of tense and number across verbal inflections.

For our experiment, we sampled 3000 instances (out of 38400 combinations of arguments and verbs) for each type
, semi-automatically crafted and manually evaluated for plausibility and grammaticality.

\subsection{Dataset variations}
\label{sec:datasets}

Starting from the main datasets described above, we build several variations that will be used in the different experiments.

\paragraph{Words} From each sentence in the type I subset of the BLM dataset,  we extract the functional words and their corresponding  nouns and prepositional phrases. There are 17 functional words and phrases: \textit{he, her, him, it, she, somebody, someone, that, that one, them, these, these ones, they, this, this one, those, those ones} and 204 noun phrases.


\paragraph{Sentences} We compile parallel versions of the sentences in their lexicalized and functional word forms from the FUN and LEX subsets of the type I BLM dataset. Each sentence has associated its syntactic pattern (the syntactic version of the syntactic-semantic template shown in Figure \ref{tab:templateBLM}, e.g. {\it Pron Vpass PP PP}). From these, we sample 4000 sentences, split 80:20 between training and testing, and use 10\% of the training data for validation. 

\paragraph{BLM data} Of the thirty verbs, all instances for three of the verbs (3x100) are selected for testing. Of the instances of the other 27 verbs, 2000 are randomly sampled for training. Ten percent of the training data is dynamically selected for validation. The same 27:3 verb split is used for all Fun/Lex and type I/ type II/type III variations. All variations have 2000 instances for training, 300 for testing. We also produced a variation where the COS and OD subtasks are merged. The data is split in a similar manner for training and testing.


\section{Analyses and experiments}
\label{sec:experiments}

We aim to determine whether language models encode the lexical abstraction property of pronouns and adverbs relative to nouns and noun phrases. We map this question into an analysis of the embedding space of the words and sentences, and proceed in several steps. 
%
   We investigate the relative positions of lexical and functional word embeddings, 
   when presented in similar sentential contexts. 
   We study the relative positions of the representations of two variations of sentences -- with nouns, or with functional words (Section \ref{sec:word_sent_embeddings}).
   We analyse the representation of functional and lexicalized sentences for detecting the shared syntactic structure (Section \ref{sec:structure}).
   We deploy the BLM linguistic puzzles, whose solution relies on detecting shared structure at the level of input sequence and within each sentence (Section \ref{sec:blm}).

We obtain word and sentence representations (as averaged token embeddings) from an Electra pretrained model \cite{clark2020electra}\footnote{{\it google/electra-base-discriminator}}. We choose Electra because it has been shown to perform better than models from the BERT family on the Holmes benchmark\footnote{The HOLMES benchmark leaderboard: \url{https://holmes-leaderboard.streamlit.app/}. At the time of writing, the ranks were: Electra - 16, DeBERTa - 21, BERT - 41, RoBERTa - 45.}, and to also encode information about syntactic and argument structure better \cite{yi-etal-2022-probing,nastase-merlo-2024-identifiable}. 

As a first step of analysis, we analyse 2D UMAP \cite{mcinnes2018umap} and t-SNE projections of the same data \cite{hinton-roweis-2002-tsne}. t-SNE is designed to project high-dimensional data into a lower dimensional space while preserving neighbourhood information, while UMAP preserves the global structure of the data.

\subsection{Contextual word embeddings}
 \label{sec:word_sent_embeddings}
 
\begin{figure}[h]
        \centering
        \includegraphics[width=0.97\linewidth]{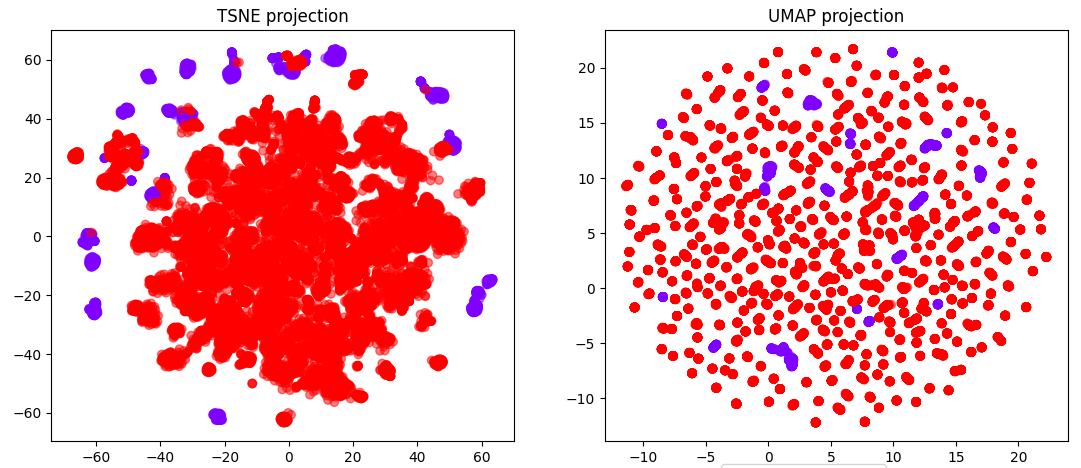}
        
        \vspace{-2mm}
        \includegraphics[width=0.2\linewidth]{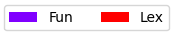}
        \vspace{-3mm}
        \caption{t-SNE and UMAP projections of the embeddings of functional words and nouns obtained from parallel contexts. Each point is a contextual embedding.
        }
        \label{fig:in-context}
    \end{figure}

We use the parallel versions of the sentences -- with content words or functional words -- to build contextualized word embeddings, and verify whether the added constraints of belonging in the same sentential contexts brings the word embeddings closer together. Each point in the plots in Figure \ref{fig:in-context} corresponds to the contextual embedding of a functional word or noun in each of the input sentences.\footnote{Word embeddings obtained for the words in isolation show the same profile (see Appendix \ref{app:embs}).}

The UMAP plot shows the embeddings of the functional words located centrally, which is congruent with their behaviour as place-holders for a wide variety of nouns. The t-SNE plot shows them separate from the other nouns, indicating that they also have specific characteristic different from nouns. These observations match the expectations of the relative position of these words in the embedding space based on the way they behave.

  \subsection{Shared structure in sentence embeddings}

  \label{sec:structure}

The analysis of the relative positions of functional words and nouns in the embedding space supports the hypothesis that the language model encodes functional words in a manner that matches their behaviour as place-holders for nouns. We deepen the exploration by checking whether the embeddings of parallel versions of sentences -- with functional words or nouns and prepositional phrases -- encode their shared syntactic-semantic information.

\begin{figure}[h]
        \centering
        \includegraphics[width=0.97\linewidth]{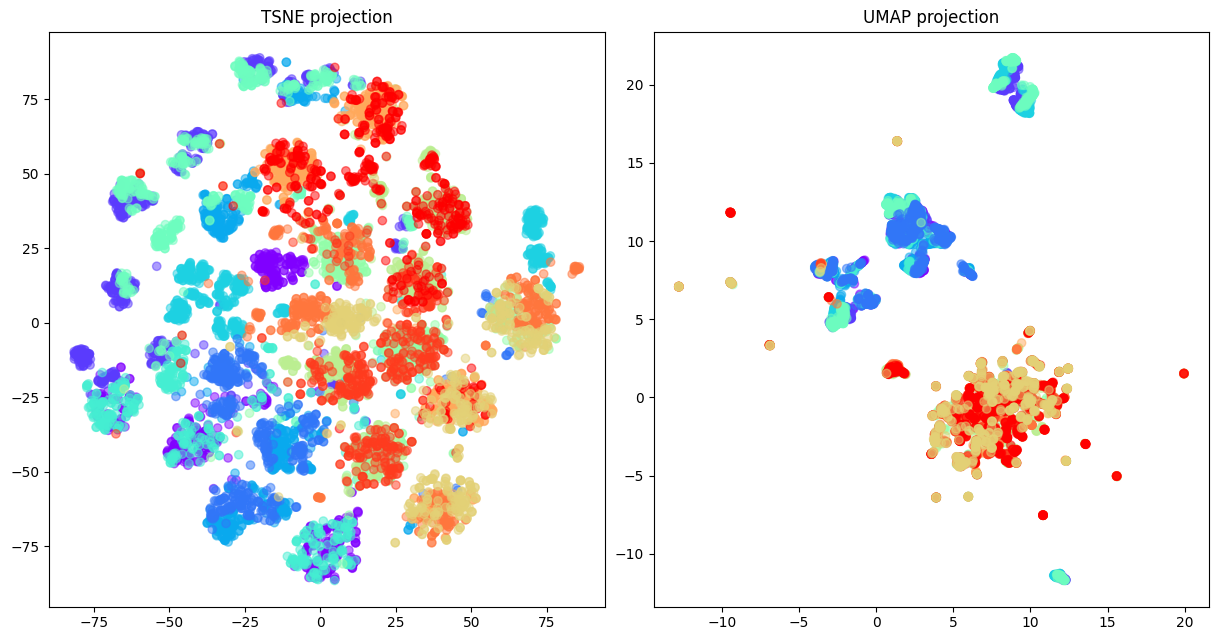}

        \includegraphics[width=0.97\linewidth]{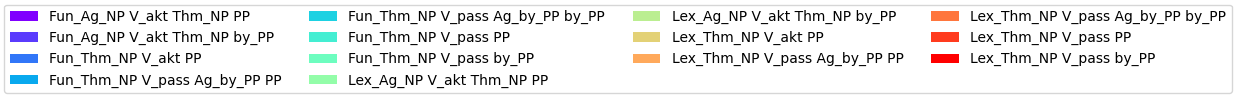}
        \vspace{-3mm}
        \caption{t-SNE and UMAP projections of sentence representations (averaged token embeddings) coloured by their syntactic pattern and the use of lexicalized or functional words.}
        \label{fig:sentence-embs}
    \end{figure}

Figure \ref{fig:sentence-embs} shows the t-SNE and UMAP projections of the representations of the two variations of each sentence. The functional and lexicalized version of the sentences occupy different regions of the embedding space, seeming to provide a negative answer to our question.

 \label{sec:sent_embeddings}

Syntactic structure and semantic roles represent complex information, which may be encoded by weighted combinations of subsets of dimensions \cite{bengio-ea2013,elhage2022toymodelssuperposition}. We mine for this information following the approach described in \citet{nastase-merlo-2024-identifiable}, which uses a variational encoder-decoder to compress sentences into representations that capture syntactic and semantic information. Sentence embeddings from Electra have size 768, and the latent layer in the used system has size 5. 
To encourage the desired information -- in this case syntactic-semantic structure -- to be encoded on the latent layer, we use contrastive learning. 

Instances consist of an input sentence $s_i$ with structure $str_i$, a positive output, i.e. a sentence $s_j\ne s_i$ that has the same structure ($str_j = str_i$), and N negative examples $s_k$ that have different structures ($str_k \ne str_i$). In our experiments we use N = 7. The structure information is used to build the dataset and obtain a deeper evaluation of the results, but is not provided to the system. We built separate datasets for Fun and Lex.

This approach enables a two-fold evaluation: (i)~in terms of performance in detecting the correct structure, by choosing the candidate answer that has the same syntactic-semantic information as the input; (ii)~in terms of the compressed representation on the latent layer, which captures these syntactic and semantic properties. 

    \begin{table}[]
        \centering
        \begin{tabular}{l|cc}
         \backslashbox{train on}{test on} & Fun & Lex \\ \hline
          Fun & 0.976 & 0.399 \\
          Lex & 0.487 & 0.914 \\ \hline
          Mixed & \textbf{0.980} & \textbf{0.916} \\
        \end{tabular}
         \caption{F1 scores on predicting the sentence with the same structure as the input, through a variational encoder-decoder system. For all experiments the system uses 2000 training instances, 10\% of which are dynamically selected in each experiment for validation. }
        \label{tab:f1-str}
    \end{table}

 \begin{figure}[h!]
        \centering
        Training on Fun
        
        \includegraphics[width=0.95\linewidth,trim={0.5cm 2.5cm 0.5cm 1cm},clip]{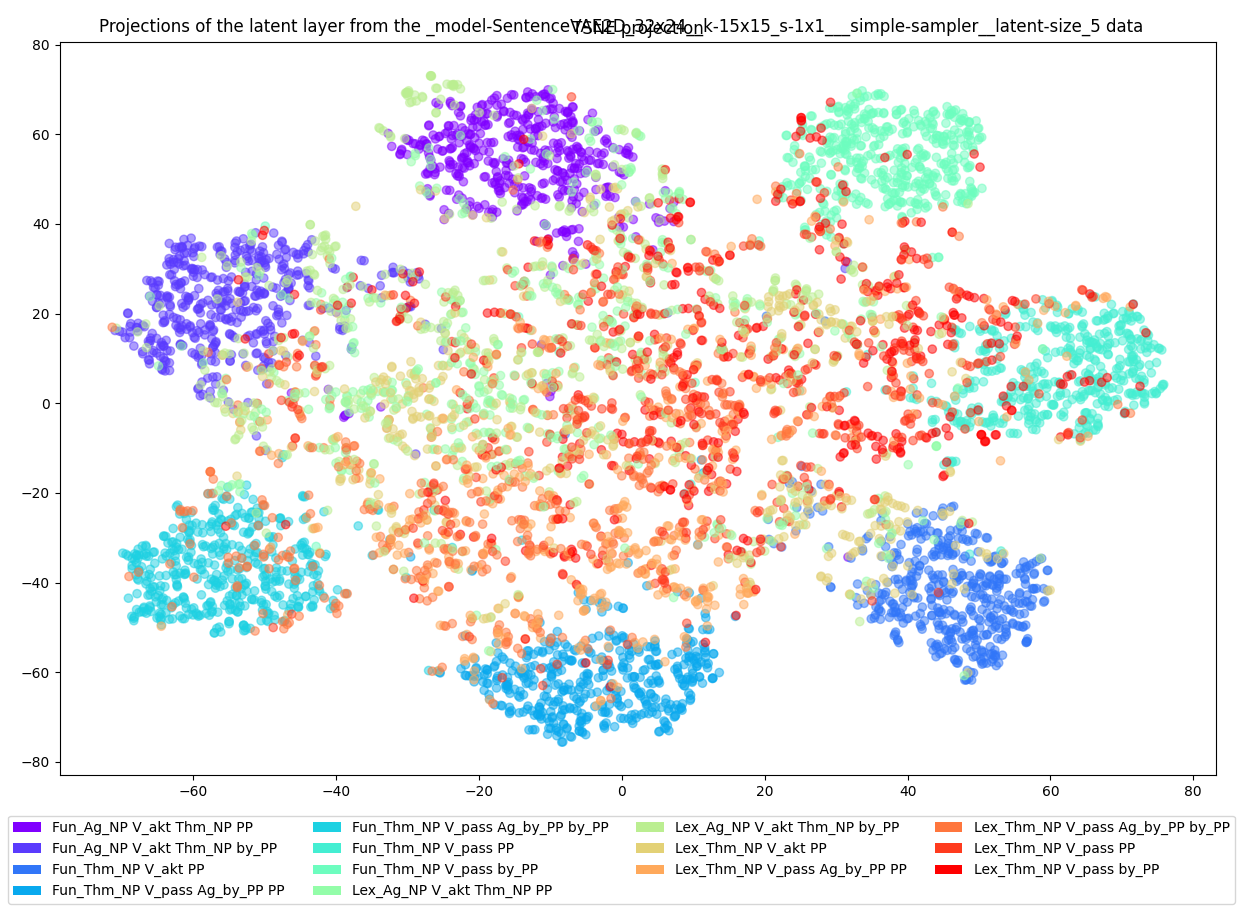}
        
        Training on Lex
        
        \includegraphics[width=0.95\linewidth,trim={0.5cm 2.5cm 0.5cm 1.15cm},clip]{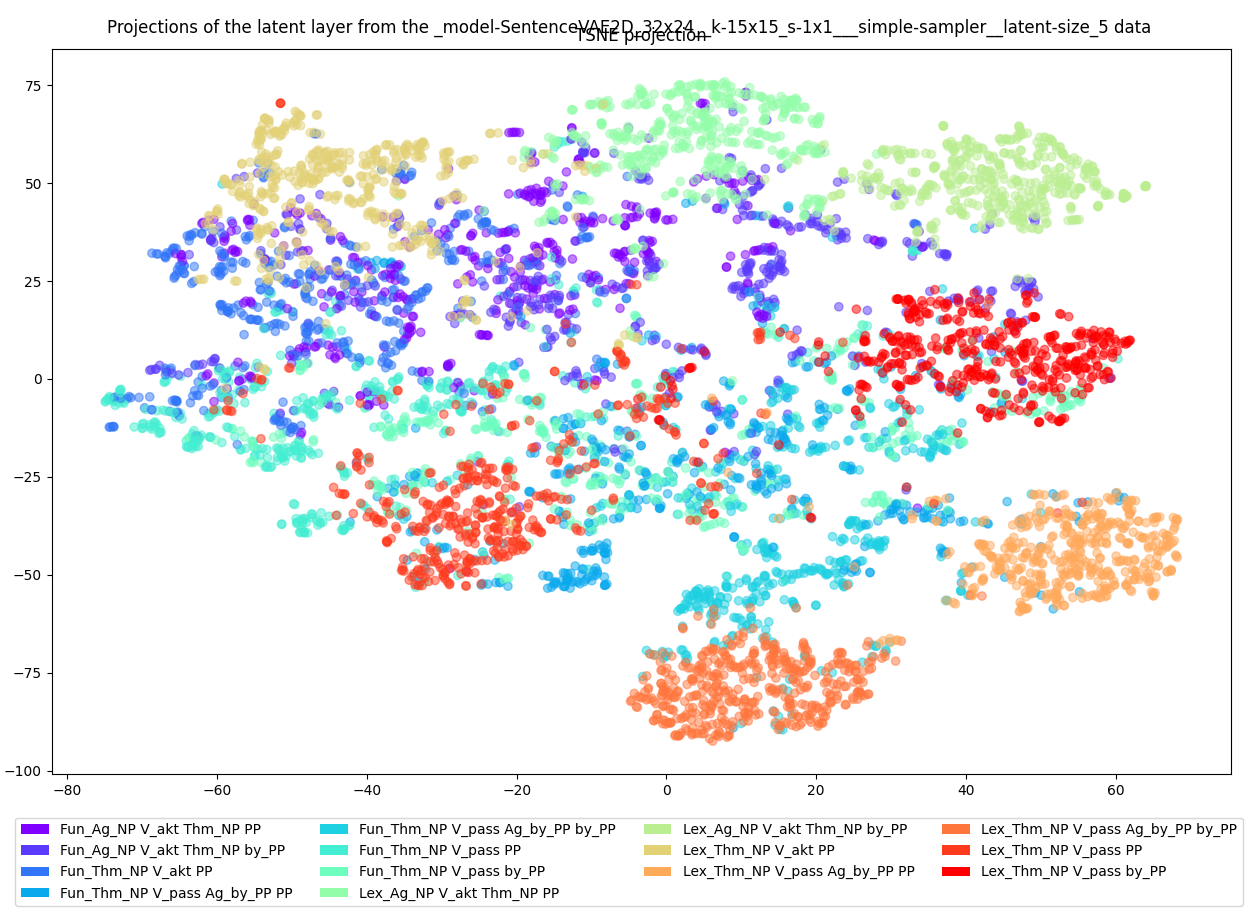}

        Training on mixed data
        
        \includegraphics[width=0.95\linewidth,trim={0.7cm 0cm 0.5cm 1cm},clip]{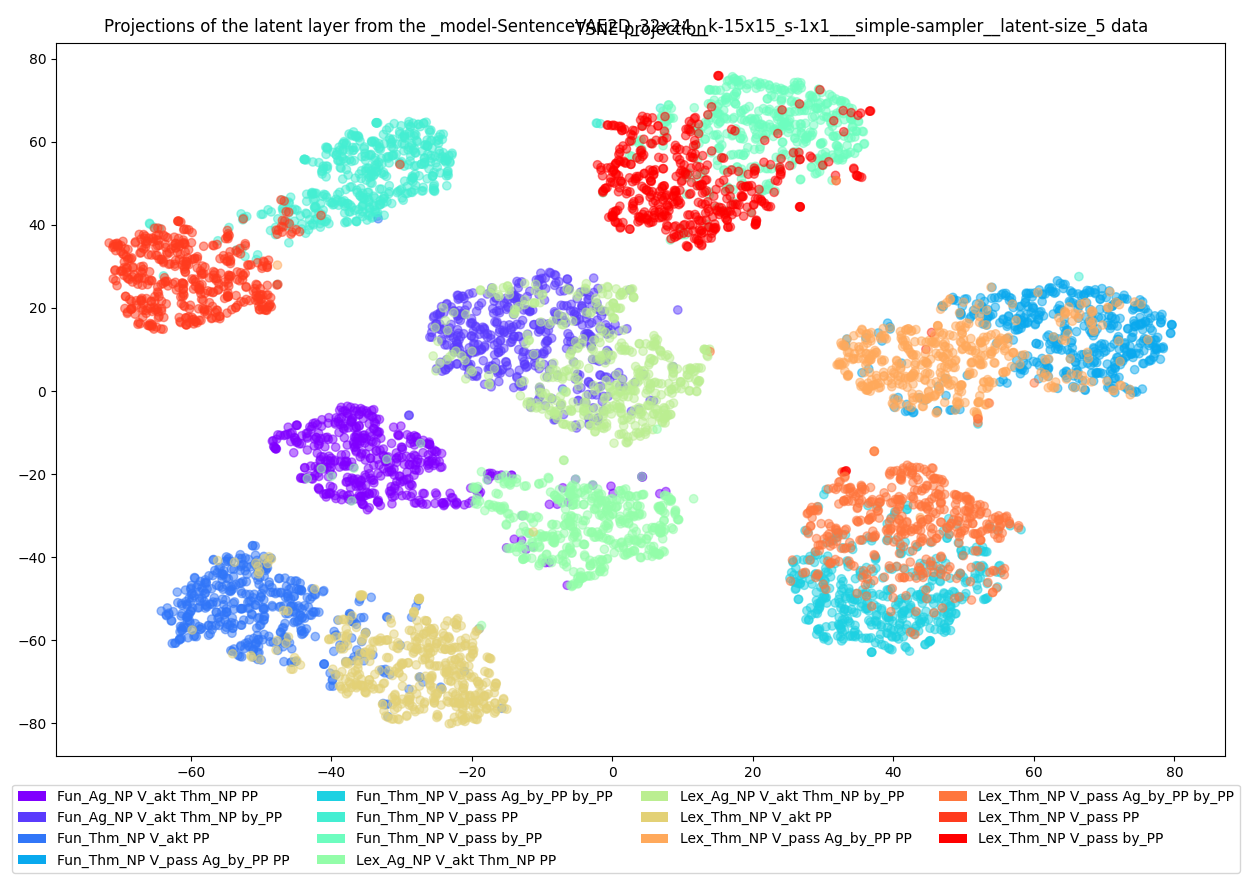}
        \caption{Latent representation analysis: t-SNE projection of vectors on the latent layer for the sentences in the training instances.}
        \label{fig:latents}
    \end{figure}

Table \ref{tab:f1-str} shows the averaged F1 scores over three experiments. We note first that training and testing on the same type (Fun or Lex) leads to high results, thus validating the experimental set-up. 

The results on test data of the same type as the training are very different from those on the test of the other type. This indicates that for each of the Fun and Lex data variations, the system discovers different clues to match two sentences with the same structure. The high results when training on the sentences with functional words may also indicate overfitting because of the repetitive vocabulary. 
Additional information comes from the analysis of the compressed representations on the latent layer, which are expected to capture the sentence structure that is shared by the functional and lexicalized data. 

The top two plots of Figure \ref{fig:latents} show the projection on the latent layer of the sentence representations with functional and content words, when trained on the sentences with functional words (top) or on the sentences with content words (middle). The plots, matching the F1 scores, show clear clusters for the data that matches the training type, but only slight separation for the data points from the other type. As discussed above, this may come from the fact that the system discovers data specific clues for each of the Fun and Lex variation. To determine whether there is a deeper, shared, layer of information in the representations of these two subsets, we train the system with a mixture of lexicalized and functionalized instances. If there is shared information, we should observe high results on both test sets when training with the mixed training data, \emph{and} overlapping clusters for the compressed representations on the latent layer. If there is no shared information, the results may be high on each test set (because separately they have been very well modelled), but the clusters of the compressed representations would be separate. 

Table \ref{tab:f1-str} shows very high results for both datasets for the mixed data training. The analysis of the representations on the latent layer, at the bottom of Figure \ref{fig:latents}, shows that the system has discovered a shared space between the sentences with functional and those with content words. What these sentences have in common is the syntactic and semantic structure, and the overlapping clusters of the compressed representations on the latent layer confirms that the system has uncovered this shared structure. The overlap between the clusters induced through joint training also supports the idea that the latent layer encodes structure, rather than simply differentiating seven amorphous classes, as there is no overlap between sentences in the Fun and Lex instances.  

    \subsection{Task solving}
    \label{sec:blm}

The previous experiments on the dataset of sentences has shown that shared structure can be detected, but it may be argued that this is a simple clustering due to other types of indicators, and the representation on the latent layer does not actually contain structure and semantic role information. We therefore use the BLM data, to investigate this deeper. The BLM task frames a linguistic phenomenon as a linguistic puzzle. Solving this puzzle relies on detecting the linguistic objects, their relevant properties, and the structure both within each sentence, and across the input sequence. 
This dataset also allows us to test generalization at several levels, because of the three levels of lexical variation. We report results on the merged COS-OD dataset (see Section \ref{sec:datasets}), as it has an added layer of complexity: the correct answer structure for a COS instance is incorrect for OD, and vice-versa. The two classes of verbs exhibit different semantic frames for the intransitive target answers (e.g., patient or agent subjects), with instances of functional and lexical elements tending to align more with either agent or patient roles.
This allows us to test whether in contexts with the two classes of verbs, functional and lexicalized phrases encode the necessary properties.



\begin{figure}
    \centering
    \includegraphics[width=0.45\textwidth]{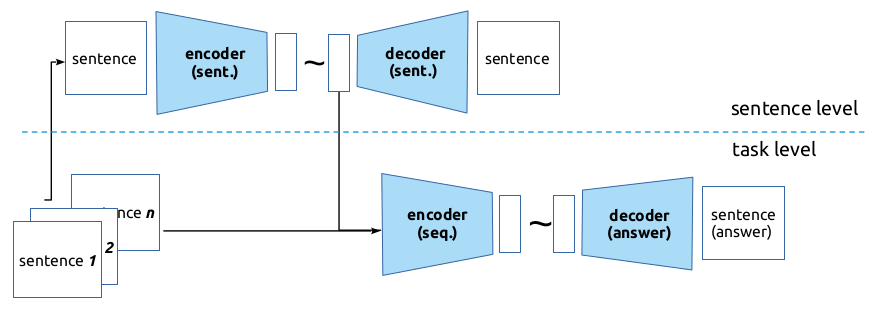}
    \caption{Two-step VAE BLM solver}
    \label{fig:2step-VAE}
\end{figure}

As the encoder-decoder set-up described above shows that the shared structural syntactic and semantic information of lexicalized and functional variations of a sentence can be detected, we incorporate this into a system that solves the BLM problem in two steps: compresses the sentence into a representation that encodes the structure relevant to the BLM puzzle, and uses these compressed representations to solve the multiple-choice puzzle \citet{nastase-merlo-2024-identifiable}. 
The two steps are encoded through interconnected variational encoder-decoders, as illustrated in Figure \ref{fig:2step-VAE}, which are trained together. The learning objective is to maximize the score of the correct answer from the candidate answer set, and minimize that of the incorrect ones. During testing, the system constructs the representation of an answer, then chooses the closest one from the given options. All potential answers consist of a verb frame filled with phrases that play specific roles (Section \ref{sec:data}). The correct one consists of the combination of phrases whose roles fit together for the given verb, while the other contain similar pieces, but which violate some semantic, syntactic (or both) rules. This set-up allows us to test whether specific elements in the sentences from the input sequence, and their semantic roles have been detected and used properly in building the correct answer.

\begin{figure}[t]
    \centering
    \includegraphics[width=\linewidth,trim={0 0.8cm 0 0}, clip]{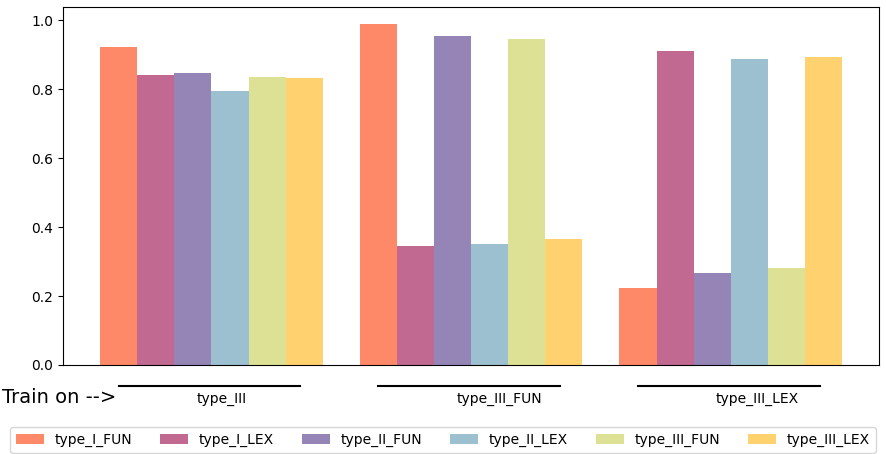}
    \caption{
    Results in terms of average F1 over three runs for solving the type III (maximal lexical variation) on merged COS and OD BLM tasks. Joint training vs. separate training. }
    \label{fig:joint_training}
\end{figure}

Figure \ref{fig:joint_training} shows the F1 results (as averages over three runs) of joint vs. separate training for the merged COS+OD BLM task. The results are for type III data, with maximum lexical variation. The complete results are in Tables \ref{tab:results_COS-OD} in the appendix. 

Processing separately datasets of sentences with and without functional words leads to high results within each task, validating the experimental set-up, but leads to low results when testing across tasks. This shows, as in the analysis of the sentences datasets, that for each of the Fun and Lex subsets, the systems discovers and exploits different regularities in the training data. Using a mixed training dataset, instead, encourages all systems to find a shared feature space. This shows that the language model encodes the syntactic-semantic structure of functional and lexicalized parallel sentences in a similar manner, but in a more complex manner that is not apparent from shallow representation similarity analyses.

\section{Discussion}

The primary goal of this paper is to investigate if LLMs encode lexical abstraction. We have translated this question into properties of words and sentence representations in the embedding space. We have analysed the relative positions of functional and content words in the embedding space, when considered in isolation, or in (syntactic and semantic) structurally parallel sentences, and we have investigated the kind of information encoded in sentence embeddings. 

\textbf{Embeddings of functional words and nouns are mingled in the embedding space, but functional words are also distinguishable from the nouns.} This result aligns with the functional words behaviour as place-holders for a wide variety of nouns and in many different contexts, but also having specific characteristics. \footnote{It is interesting to remark, in this respect, that the semantic literature also contains proposals suggesting that pronominal forms are not place-holders, but are better considered as equivalent to noun phrase (NP) descriptions, where they refer to a less abstract, fuller expression in context, in relevant environments \citep{elbourne2002situations,lewis2022descriptions}.}  

\textbf{Sentence embeddings of parallel sentence variations occupy different regions of the embedding space.} This result seems to show that functional and lexical sentence variations do not share much information, as their distance in the embedding space indicates low similarity. 

\textbf{We can detect information about the shared syntactic structure in the embeddings of the functional and lexical variations of the same sentences.}  Our follow-up experiments uncover information about shared information in Fun and Lex variation of sentences, and of a more complex linguistic puzzle. Training on only one type of data does not reveal the shared syntactic structure and semantic roles, reflecting the shallow differences noted in the sentence embedding space. Training on mixed data, however, leads to high results on both dataset variations \emph{and} overlapping clustered representations in the latent space.

Contrary to our conclusion that the system has discovered a shared space based on the abstraction of nouns, one might argue that the shared space we find is due only to the shared verb, or shared lexical material between the train and test partitions. Had that been the case, the cross-testing results, when training on separate data types, would have been closer to the  results on mixed data, given that the verb is not replaced by a functional category and it remains the same across all types of data and sentences. This argument is  especially true for the type III subset of the BLM task, which has maximal lexical variation.  

We think instead that the results indicate that the model trained on the functional data, which has a very small and consistent vocabulary, relies on shallower features, while the model learned on the lexicalized data is more robust, but not sufficiently abstract. Training the system with mixed data leads not only to a model that performs very well on both data variations, but all sentences are projected into the same compressed embedding space, establishing the necessary links between nominal expressions and their functional equivalents that support abstraction and generalisation.

      \section{Related work}

A generalization taxonomy based on an extensive analysis of publications in NLP that deal with the topic of generalization is proposed in \citet{Hupkes2023}. They distinguish five main dimensions for generalization analysis: motivation (concerning the higher-level aims of the model), generalization type (the properties of language or domain or model the model is intended to capture), shift type (the kind of differences between training and testing data distributions), shift source (the source of the difference in data distributions) and shift locus (where in the pipeline does the shift in data distributions occurs). This analysis reflects the focus in the NLP community on the model, and its properties from a machine learning point of view. 

Language has its own generalization and abstraction dimensions, which could be at the lexical level \cite{regneri-etal-2024-detecting,sukumaran-etal-2024-investigating}, concern verb frames \cite{wilson-etal-2023-abstract,yi-etal-2022-probing}, grammar \cite{kim-smolensky-2021-testing} or a combination of these \cite{wang-etal-2024-abspyramid}. The results of such investigations do not reveal a clear picture. While \citet{kim-smolensky-2021-testing} observe a limited degree of generalization based on grammatical categories, they note that the results may not have been driven by abstraction. \citet{yi-etal-2022-probing} show that both verb and sentence representations encode information about a verb's alternation class, but the linguistic generalization within the verb argument structure is limited, as models fail on unseen contexts. In experiments on an entailment graph  that contains abstract concepts entailed by components of events (nouns, verbs, the event as a whole), \citet{wang-etal-2024-abspyramid} show that the LLMs have difficulty understanding abstract knowledge, but they can be improved with fine-tuning.


Structural priming is used in \citet{michaelov-etal-2023-structural} to investigate the degree of grammatical abstraction in LLMs for three verb alternations: active/passive, dative alternation and two forms of possessive. In monolingual and cross-lingual settings, they find evidence for abstract grammatical representations of these phenomena. 



Close to the topic of this paper, \citet{regneri-etal-2024-detecting} investigate whether hyponymy is encoded in the transformer by analysing the attention matrices when presented with hyponymous noun pairs. In our work, instead, we have analysed the output of a pretrained language model, and whether the word and sentence embeddings it produces encode particular linguistic information that would allow us to establish a parallel between lexicalized and abstract expressions of a sentence.

All this work shows an unclear picture of sentence embeddings, and the information -- and its degree of abstractness -- it encodes.
Our work provides further linguistically-oriented evidence to clarify the relation between embeddings, abstraction and generalisation.

\section{Conclusions}

Our study contributes to the discussion of generalization in language models, and in particular studies linguistic generalization. 
We translate the question of whether language models capture lexical abstraction into tests of word and sentence embeddings in the embedding space. We show that the linguistic behaviour of functional words as place-holders for nouns is reflected in the relative positions of their corresponding embeddings. We show that despite the apparent lack of similarity between functional and lexicalized versions of sentences -- as shown by their distinct positions in the embedding space -- a deeper analysis reveals shared syntactic and semantic information. These conclusions are further reinforced by the results on a problem solving task, the BLM multi-choice problem, whose solution relies on the proper detection of linguistic objects and their relations.


\section{Limitations}





\paragraph{Synthetic dataset} We use a synthetic dataset, for controlled experimentation, which primarily consists of simple sentence structures.  The dataset, then,  may not fully capture the complexity of language. Future extensions will include many more structures and variations.  Another limitation is the all-or-nothing pronominalisation of sentences, where each sentence is either fully categorized into a predefined functional element or not. Future work will have to modulate the amount of pronominalisation and study different patterns of interactions between nominal expressions and their pronominal equivalent. Moreover, at the moment, we do not have comparable results with a human experiment, which could shed light on more human-like abstraction processes. Finally, this study relies exclusively on English data. While many pronominal systems are structured like the one of English, many other pronominal systems exist. \citet{nastase-etal-2025-multilingual} discusses similar problems for Italian, while future studies should add a cross-linguistic dimension.

\paragraph{Using encoder transformer models} Previous reviewers have commented on the fact that we do not use multiple LLMs for this task, and in particular no generative models. We have justified our choice of the model in Section \ref{sec:experiments}, as Electra is one of the best performing encoder models, and also outperforms XLNet \cite{yang2019xlnet} and MPNet \cite{song2020mpnet} on GLUE tasks. The embedding space of different models is different, and we chose to study the embedding space of one of the best performing encoder-based models. For our study we require word embeddings as well as sentence embeddings. While there are ways to elicit approximations of sentence embeddings through word-definition like prompts \cite{jiang-etal-2024-scaling,zhang2024simple}, these are only approximations and not direct representations of a target sentence, which is important particularly when the sentences in our data have specific linguistic and grammatical properties, and do not have obvious one word equivalents.

\bibliography{custom,references}

\newpage
\appendix

\onecolumn

\section{Data}

\begin{table}[h]
    \centering
    \renewcommand{\arraystretch}{1.2}
    \begin{tabular}{l p{15cm}} 
        \hline 
        \textbf{Class} & \textbf{Verb} \\
        \hline 
        COS & \textit{bake}, \textit{bend}, \textit{blacken}, \textit{break}, \textit{brighten}, \textit{caramelize}, \textit{chip}, \textit{close}, \textit{corrode}, \textit{crinkle}, \textit{defrost}, \textit{empty}, \textit{expand}, \textit{fry}, \textit{harden}, \textit{harmonize}, \textit{heat}, \textit{improve}, \textit{increase}, \textit{intensify}, \textit{melt}, \textit{open}, \textit{propagate}, \textit{purify}, \textit{sharpen}, \textit{shrink}, \textit{sweeten}, \textit{tear}, \textit{whiten}, \textit{widen}. \\
        OD & \textit{clean}, \textit{cook}, \textit{draw}, \textit{drink}, \textit{eat}, \textit{fish}, \textit{hum}, \textit{iron}, \textit{knead}, \textit{knit}, \textit{mend}, \textit{milk}, \textit{nurse}, \textit{paint}, \textit{play}, \textit{plow}, \textit{polish}, \textit{read}, \textit{recite}, \textit{sculpt}, \textit{sew}, \textit{sing}, \textit{sow}, \textit{study}, \textit{sweep}, \textit{teach}, \textit{wash}, \textit{weave}, \textit{whittle}, \textit{write}. \\
        \hline
    \end{tabular}
    \caption{Verbs categorized by class}
    \label{tab:verbs-app}
\end{table}

\begin{figure*}[!h]
\footnotesize
\setlength{\tabcolsep}{5pt} 
\begin{tabular}{|p{0.1cm}|p{8.5cm}|} 
\hline
\multicolumn{2}{|c|}{\sc COSFun - Context} \\
\hline						
1 & She broke it with this \\
2 & She broke it by those there \\
3 & It was broken by her with this \\
4 & It was broken by her by those there \\
5 & It was broken with this \\
6 & It was broken by those there \\
7 & It broke with this \\
? & ??? \\ \hline 
\end{tabular} 
\begin{tabular}{|p{0.1cm}|p{6.5cm}|} 
\hline
\multicolumn{2}{|c|}{\sc COSFun - Answers} \\
\hline
1 & \textbf{It broke by those there} \\
2 & She broke by those there \\
3 & It was broken by her \\
4 & She was broken by it\\
5 & It broke her \\
6 & She broke it \\
7 & It broke by her \\
8 & She broke by it\\
\hline
\end{tabular}
\\ 
\begin{tabular}{|p{0.1cm}|p{8.5cm}|} 
\hline
\multicolumn{2}{|c|}{\sc COSLex - Context} \\
\hline
1 & The archaeologist broke a vase in the lab  \\
2 & The archaeologist broke a vase by mistake\\
3 & The vase was broken by the archaeologist in the lab\\
4 & The vase was broken by the archaeologist by mistake \\
5 & The vase was broken in the lab\\
6 & The vase was broken by mistake\\
7 & The vase broke in the lab\\
? & ??? \\ 
\hline 
\end{tabular} 
\begin{tabular}{|p{0.1cm}|p{6.5cm}|} 
\hline
\multicolumn{2}{|c|}{\sc COSLex - Answers} \\
\hline
1 & \textbf{The vase broke by mistake} \\
2 & The archaeologist broke by mistake \\
3 & The vase was broken by the archaeologist \\
4 & The archaeologist was broken by the vase\\
5 & The vase broke the archaeologist \\
6 & The archaeologist broke the vase \\
7 & The vase broke by the archaeologist \\
8 & The archaeologist broke by the vase \\
\hline
\end{tabular}
\caption{Examples of \textsc{Fun} and \textsc{Lex} for the English verb \textit{break}, one of the verbs belonging to COS class.}
\label{tab:TypeI-COSfunlex}
\end{figure*}

\begin{figure*}[!h]
\footnotesize
\setlength{\tabcolsep}{5pt} 
\begin{tabular}{|p{0.1cm}|p{8.5cm}|} 
\hline
\multicolumn{2}{|c|}{\sc ODLex - Context} \\
\hline						
1 & They paint it with this \\
2 & They paint it by that \\
3 & It was painted by them with this \\
4 & It was painted by them by that \\
5 & It was painted with this \\
6 & It was painted by that \\
7 & They painted with this \\
? & ??? \\ \hline 
\end{tabular} 
\begin{tabular}{|p{0.1cm}|p{6.5cm}|} 
\hline
\multicolumn{2}{|c|}{\sc ODFun - Answers} \\
\hline
1 & It painted by that \\
2 & \textbf{They painted by that}\\
3 & It was painted by them  \\
4 & They were painted by it\\
5 & It painted them \\
6 & They painted it \\
7 & It painted by them \\
8 & They painted by it\\
\hline
\end{tabular}
\\ 
\begin{tabular}{|p{0.1cm}|p{8.5cm}|} 
\hline
\multicolumn{2}{|c|}{\sc COSLex - Context} \\
\hline
1 & These artists paint a portrait with a   brush \\
2 &  These artists paint a portrait by the lake\\
3 & A portrait was painted by these artists with a brush\\
4 & A portrait was painted by these artists by the lake \\
5 & A portrait was painted with a brush \\
6 & A portrait was painted by the lake\\
7 & These artists painted with a brush \\
? & ??? \\ 
\hline 
\end{tabular} 
\begin{tabular}{|p{0.1cm}|p{6.5cm}|} 
\hline
\multicolumn{2}{|c|}{\sc COSLex - Answers} \\
\hline
1 & A portrait painted by the lake \\
2 & \textbf{These artists painted by the lake} \\
3 & A portrait was painted by the artists \\
4 & These artists were painted by a portrait\\
5 & A portrait painted these artists \\
6 & These artists painted a portrait \\
7 & A portrait painted by these artists \\
8 & These artists painted by a portrait \\
\hline
\end{tabular}
\caption{Examples of Type\_I \textsc{Fun} and \textsc{Lex} data for the English verb \textit{paint}, one of the verbs belonging to OD class}
\label{tab:TypeI-ODfunlex}
\end{figure*}

\section{Stand-alone embeddings}
\label{app:embs}

\begin{figure}[h]
        \centering
        \includegraphics[width=0.95\linewidth]{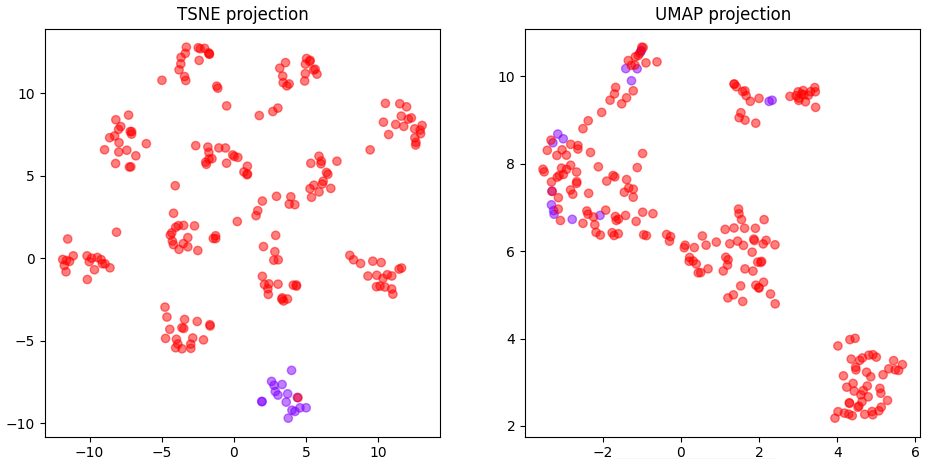}
        \includegraphics[width=0.2\linewidth]{figs/fun_vs_lex_legend.png}
        \vspace{-3mm}
        \caption{t-SNE and UMAP projections of the embeddings of functional words and nouns, without a sentential context. 
        }
        \label{fig:stand-alone}
    \end{figure}

    Figure \ref{fig:stand-alone} shows the t-SNE projection of the word embeddings (as averages over the respective token embeddings) for the functional words and noun phrases in our data, obtained in isolation (when presented to the pretrained model alone).    Functional words appear isolated in this space, which indicates that the shared information between the functional elements and the nouns they can replace, should there by any, is not to be found at a shallow level.

\newpage
\section{BLM task results}
\label{app:blm}

The experiments were run on an HP PAIR Workstation Z4 G4 MT, with an Intel Xeon W-2255 processor, 64G RAM, and a MSI GeForce RTX 3090 VENTUS 3X OC 24G GDDR6X GPU. The systems were trained for 300 epochs, and the results reported are F1 averages (standard deviation) over three runs. The training data for all experiments was 2000 instances. For the joint-training set-up, the data was split evenly across the variations.

\begin{figure}[t]
    \centering
    COS \\
    \includegraphics[width=0.8\linewidth,trim={0 0.8cm 0 0}, clip]{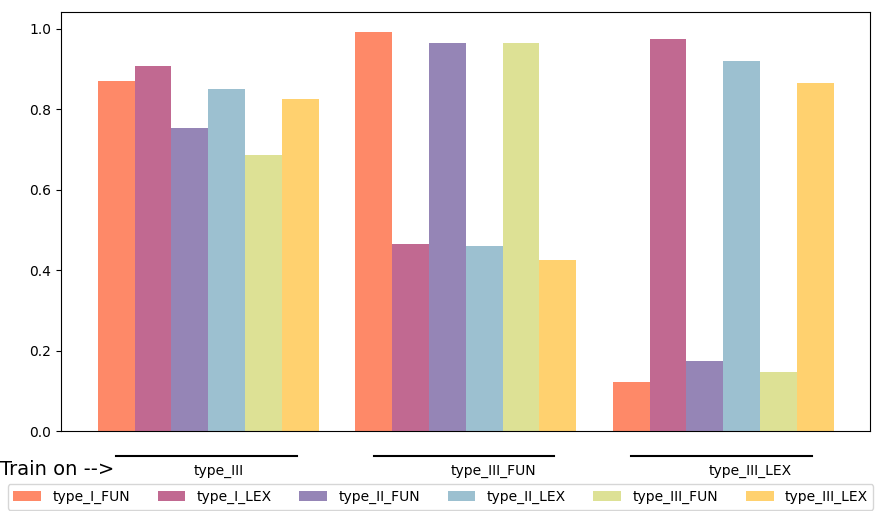}
    ~\\ 
    \vspace{3mm}
    OD \\
    \includegraphics[width=0.8\linewidth]{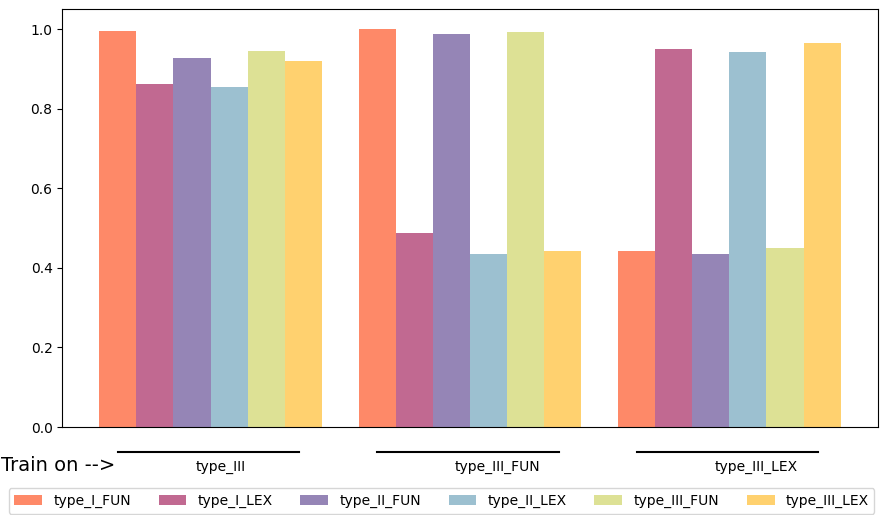}

    \caption{
    Results in terms of average F1 over three runs for solving the type III (maximal lexical variation) COS and OD BLM tasks for three models. Joint training vs. separate training. }
    \label{fig:separate_training}
\end{figure}

\begin{table}[h]
\centering
 \begin{tabular}{lccc}
 \textbf{test on} & \multicolumn{3}{c}{\textbf{train on}} \\ \hline 
  & \multicolumn{3}{c}{Joint training} \\
               & type\_I  & type\_II & type\_III \\ \hline
  type\_I\_Fun & 0.811 (0.039)  & 0.892 (0.013) &  0.922 (0.010) \\
  type\_I\_Lex & 0.599 (0.029) & 0.579 (0.014) & 0.842 (0.018) \\ \cline{2-4}
  type\_II\_Fun & 0.662 (0.021) & 0.803 (0.003) & 0.849 (0.018) \\
  type\_II\_Lex & 0.542 (0.028)  & 0.532 (0.007) &  0.794 (0.012) \\ \cline{2-4}
  type\_III\_Fun & 0.686 (0.021) &  0.816 (0.015) &  0.837 (0.015) \\
  type\_III\_Lex & 0.691 (0.007) & 0.765 (0.010) & 0.833 (0.012) \\ \hline 
 & \multicolumn{3}{c}{Training on Fun} \\ 
       & type\_I\_Fun  & type\_II\_Fun & type\_III\_Fun \\ \hline
 type\_I\_Fun & 0.948 (0.005)  & 0.947 (0.013) & 0.989 (0.004) \\
 type\_I\_Lex & 0.416 (0.023) & 0.396 (0.004) &  0.346 (0.008) \\ \cline{2-4}
 type\_II\_Fun & 0.854 (0.010) & 0.914 (0.009) & 0.955 (0.005) \\
 type\_II\_Lex & 0.398 (0.018) & 0.365 (0.012) & 0.351  (0.003) \\ \cline{2-4}
 type\_III\_Fun & 0.852 (0.008) & 0.927 (0.007) & 0.947 (0.005) \\
 type\_III\_Lex & 0.382 (0.018)  & 0.415 (0.024) & 0.366 (0.008)\\ \hline
 
 & \multicolumn{3}{c}{Trainig on Lex} \\
       & type\_I\_Lex  & type\_II\_Lex & type\_III\_Lex \\ \hline
 type\_I\_Fun & 0.236 (0.031)  & 0.235 (0.018)  & 0.224 (0.018) \\
 type\_I\_Lex & 0.574 (0.009) & 0.672 (0.010)  & 0.913  (0.006) \\  \cline{2-4}
 type\_II\_Fun & 0.261 (0.017)  & 0.283 (0.006) & 0.267  (0.008) \\
 type\_II\_Lex & 0.549 (0.008) & 0.639 (0.018) & 0.889 (0.008)  \\  \cline{2-4}
 type\_III\_Fun & 0.286 (0.024) & 0.288 (0.015) & 0.282 (0.006) \\
 type\_III\_Lex & 0.669 (0.013) & 0.828 (0.007) & 0.893 (0.002) \\ \hline 
 \end{tabular}
 \caption{COS and OD merged tasks: Results as averaged F1 (std) over three runs, for three training set-ups: joint training (training using both Fun and Lex instances), training on Fun instances, training on Lex instances. For all set-ups we use 2000 training instances. For the joint training these are evenly split between Fun and Lex.}
 \label{tab:results_COS-OD}
\end{table}

\begin{table}[h]
\centering
 \begin{tabular}{lccc}
 \textbf{test on} & \multicolumn{3}{c}{\textbf{train on}} \\ \hline 
  & \multicolumn{3}{c}{Joint training} \\
               & type\_I  & type\_II & type\_III \\ \hline
  type\_I\_Fun & 0.866 (0.010)  & 0.893 (0.010) &  0.871 (0.010) \\
  type\_I\_Lex & 0.861 (0.002) & 0.921 (0.013) & 0.908 (0.009) \\ \cline{2-4}
  type\_II\_Fun & 0.723 (0.020) & 0.812 (0.016) & 0.754 (0.010) \\
  type\_II\_Lex & 0.808 (0.006)  & 0.873 (0) &  0.851 (0.002) \\ \cline{2-4}
  type\_III\_Fun & 0.680 (0.009) &  0.770 (0.024) &  0.686 (0.010) \\
  type\_III\_Lex & 0.790 (0.003) & 0.800 (0.021) & 0.826 (0.011) \\ \hline 
 & \multicolumn{3}{c}{Training on Fun} \\ 
       & type\_I\_Fun  & type\_II\_Fun & type\_III\_Fun \\ \hline
 type\_I\_Fun & 0.981 (0.006)  & 0.973 (0.003) & 0.992 (0.004) \\
 type\_I\_Lex & 0.404 (0.023) & 0.433 (0.014) &  0.464 (0.006) \\ \cline{2-4}
 type\_II\_Fun & 0.927 (0.018) & 0.933 (0.011) & 0.964 (0.009) \\
 type\_II\_Lex & 0.396 (0.014) & 0.427 (0.012) & 0.459  (0.007) \\ \cline{2-4}
 type\_III\_Fun & 0.898 (0.012) & 0.933 (0.005) & 0.963 (0.003) \\
 type\_III\_Lex & 0.388 (0.024)  & 0.414 (0.026) & 0.426 (0.011)\\ \hline
 
 & \multicolumn{3}{c}{Trainig on Lex} \\
       & type\_I\_Lex  & type\_II\_Lex & type\_III\_Lex \\ \hline
 type\_I\_Fun & 0.160 (0.017)  & 0.117 (0.005)  & 0.122 (0.019) \\
 type\_I\_Lex & 0.986 (0.006) & 0.960 (0.007)  & 0.973  (0.005) \\  \cline{2-4}
 type\_II\_Fun & 0.240 (0.005)  & 0.132 (0.011) & 0.174  (0.010) \\
 type\_II\_Lex & 0.877 (0.008) & 0.908 (0.008) & 0.920 (0.007)  \\  \cline{2-4}
 type\_III\_Fun & 0.231 (0.007) & 0.150 (0.010) & 0.147 (0.005) \\
 type\_III\_Lex & 0.809 (0.006) & 0.846 (0.010) & 0.864 (0.004) \\ \hline 
 \end{tabular}
 \caption{BLM-COS: Results as averaged F1 (std) over three runs, for three training set-ups: joint training (training using both Fun and Lex instances), training on Fun instances, training on Lex instances. For all set-ups we use 2000 training instances. For the joint training these are evenly split between Fun and Lex.}
 \label{tab:results_COS}
\end{table}

\begin{table}[h]
\centering
 \begin{tabular}{lccc}
 \textbf{test on} & \multicolumn{3}{c}{\textbf{train on}} \\ \hline 
 & \multicolumn{3}{c}{Joint training} \\ 
               & type\_I  & type\_II & type\_III \\ \hline
type\_I\_Fun & 0.979 (0.006)  & 0.970 (0.007)  & 0.994 (0.003) \\
type\_I\_Lex & 0.683 (0.014) & 0.764 (0.016) & 0.861 (0.014) \\ \cline{2-4}
type\_II\_Fun & 0.854 (0.023) & 0.898 (0.010) & 0.927 (0.005) \\
type\_II\_Lex & 0.603 (0.021) & 0.701 (0.023) & 0.854 (0.006) \\ \cline{2-4}
type\_III\_Fun & 0.886 (0.020) & 0.947 (0.020) & 0.944 (0.004) \\
type\_III\_Lex & 0.849  (0.011) & 0.882 (0.016) & 0.920 (0.007) \\ \hline
 & \multicolumn{3}{c}{Train on Fun} \\
 & type\_I\_Fun  & type\_II\_Fun & type\_III\_Fun \\ \hline
type\_I\_Fun & 1.000 (0)  & 1.000 (0) & 1.000 (0) \\
type\_I\_Lex & 0.474 (0.011) & 0.549 (0.038) & 0.489 (0.039) \\ \cline{2-4}
type\_II\_Fun & 0.923 (0.010) & 0.962 (0.008) & 0.989 (0.007)  \\
type\_II\_Lex & 0.431 (0.036) & 0.476 (0.026) & 0.436 (0.022) \\ \cline{2-4}
type\_III\_Fun & 0.942 (0.014) & 0.980 (0.005) & 0.993 (0.003) \\
type\_III\_Lex & 0.365 (0.015) & 0.487 (0.017) & 0.443  (0.012) \\ \hline
& \multicolumn{3}{c}{Training on Lex} \\
      & type\_I\_Lex  & type\_II\_Lex & type\_III\_Lex \\ \hline  
type\_I\_Fun & 0.489 (0.037) & 0.480 (0.007)  & 0.442 (0.020) \\
type\_I\_Lex & 0.746 (0.007) & 0.796 (0.015) & 0.951 (0.013) \\ \cline{2-4}
type\_II\_Fun & 0.43 (0.005) & 0.467 (0.005) & 0.436 (0.018) \\
type\_II\_Lex & 0.682 (0.026) & 0.712 (0.010) & 0.943 (0.014) \\ \cline{2-4}
type\_III\_Fun & 0.501 (0.008) & 0.466 (0.011) & 0.451 (0.014) \\
type\_III\_Lex & 0.876 (0.017) & 0.899 (0.008) & 0.964  (0.009) \\ \hline
\end{tabular}
 \caption{BLM-OD: Results as averaged F1 (std) over three runs, for three training set-ups: joint training (training using both Fun and Lex instances), training on Fun instances, training on Lex instances. For all set-ups we use 2000 training instances. For the joint training these are evenly split between Fun and Lex.}
 \label{tab:results_OD}
\end{table}

\end{document}